\documentclass[11pt]{article}
\usepackage{nodalida2025}
\usepackage{times}
\usepackage{url}
\usepackage{booktabs}
\usepackage{latexsym}
\usepackage{graphicx}

\usepackage[T1]{fontenc}
\usepackage[utf8]{inputenc}

\usepackage{hyperref}

\newcommand{\sortas}[1]{}

\title{GliLem: Leveraging GliNER for Contextualized Lemmatization in Estonian}

\author{Aleksei Dorkin \\
  Institute of Computer Science \\ University of Tartu \\
  {\tt aleksei.dorkin@ut.ee} \\\And
  Kairit Sirts \\
  Institute of Computer Science \\ University of Tartu \\
  {\tt kairit.sirts@ut.ee} \\
  }

\date{}

\begin{document}
\maketitle
\begin{abstract}
We present GliLem---a novel hybrid lemmatization system for Estonian that enhances the highly accurate rule-based morphological analyzer Vabamorf with an external disambiguation module based on GliNER---an open vocabulary NER model that is able to match text spans with text labels in natural language. We leverage the flexibility of a pre-trained GliNER model to improve the lemmatization accuracy of Vabamorf by 10\% compared to its original disambiguation module and achieve an improvement over the token classification-based baseline. To measure the impact of improvements in lemmatization accuracy on the information retrieval downstream task, we first created an information retrieval dataset for Estonian by automatically translating the DBpedia-Entity dataset from English. We benchmark several token normalization approaches, including lemmatization, on the created dataset using the BM25 algorithm. We observe a substantial improvement in IR metrics when using lemmatization over simplistic stemming. 
The benefits of improving lemma disambiguation accuracy manifest in small but consistent improvement in the IR recall measure, especially in the setting of high~k.\footnote{A demo of the system is available at \url{https://huggingface.co/spaces/adorkin/GliLem}}
\end{abstract}

\section{Introduction}

Lemmatization plays an important role in natural language processing by reducing words to their base or dictionary forms, known as lemmas. This process is especially crucial for morphologically rich languages such as Estonian, where words can exhibit a multitude of inflected forms. Effective lemmatization enhances various downstream NLP tasks, including information retrieval based on lexical search and text analysis. Although dense vector retrieval is gaining traction in information retrieval, lexical search methods remain highly relevant, particularly in modern hybrid systems. Lexical search excels as a first-stage retriever due to its efficiency with inverted indices, and provides reliable exact term matching that dense retrievers may miss~\cite{gao2021complement}. Recent research demonstrates that lexical and dense retrieval are complementary, lexical matching providing a strong foundation for precise word-level matches, while dense retrieval captures semantic relationships and handles vocabulary mismatches. The complementary nature of these approaches has led to state-of-the-art hybrid systems that outperform either method alone~\cite{lee2023complementarity}.

Vabamorf~\cite{kaalep2001complete} is a rule-based morphological analyzer for the Estonian language. It provides one or more morphological analysis (including lemma) candidates for each token in a text, where the token can be a word or a punctuation mark. 
The Vabamorf's analyzer functionality aims to generate all possible morphological analysis and lemma candidates for each word, regardless of its context.
However, in order to find the appropriate analysis together with the lemma in a particular textual context, the analyzer output needs to be disambiguated.
Vabamorf employs a built-in Hidden Markov Model (HMM) based disambiguator that can only look at the word's immediate context to rank the analysis candidates by their likelihood scores. 
Thus, despite its high precision in generating lemma candidates, Vabamorf's ability to disambiguate these candidates in context is limited due to its weak representational power.

Previously, \citet{dorkin-sirts-2023-comparison} have shown that, when evaluated on the Estonian Universal Dependencies corpus, Vabamorf's disambiguation abilities reach to ca 89\% for lemmatization.
However, when evaluated in the oracle mode, where a prediction is considered correct if the true lemma appears among the candidates, it achieves an accuracy above 99\%.\footnote{For instance, if Vabamorf outputs three distinct lemma candidates for a given token, the oracle considers the prediction correct if one of these candidates is correct. This approach is unusable in a practical scenario, because the predictions have to be disambiguated.}
This significant difference highlights the limitations of the Vabamorf's current disambiguator and underscores the need for improving its disambiguation component.

Recent methods to neural lemmatization generally follow two approaches: pattern-based token classification \cite{straka-2018-udpipe,straka2019udpipe} and generative modeling \cite{kanerva2018turku,kanerva2021universal}.
The pattern-based approach predicts for each word a transformation pattern that can be used to transform the word token into corresponding lemma. When built on top of contemporary BERT-based encoders, the pattern-based lemmatizer makes use of the contextual token representations directly to make the prediction.
The generative approach uses a character-based sequence-to-sequence model to generate the lemma conditioned on the word, relying on disambiguated morphological information as context. 
While both of these approaches have shown good results on Estonian \cite{dorkin-sirts-2023-comparison}, neither of them is well suited for developing a new disambiguator for Vabamorf. 
First, the pattern-based token classification approach operates with a limited pattern vocabulary extracted from a training set and cannot handle previously unseen patterns that may be output by Vabamorf.
Secondly, the generative model already assumes the presence of disambiguated morphological analyses making the disambiguation problem circular.

Recently, an open vocabulary model GliNER for Named Entity Recognition (NER) was proposed by \citet{zaratiana2023gliner} which can be used to match arbitrary text labels with input text spans. In the lemmatizer disambiguation setting, the GliNER approach can be used to match the transformation patterns extracted from Vabamorf analysis candidates to the spans of sub-word tokens making up words in the text, making it suitable for scoring a limited number of lemma candidates for each word.

Our first aim in this paper is to investigate whether GliNER method can be used to disambiguate the Vabamorf's lemma candidates. For that, we modify the GliNER implementation to predict the transformation patterns of Vabamorf's generated lemma candidates, using the Estonian Universal Dependencies corpus~\cite{11234/1-5150} for training. We find that using this approach boosts the disambiguation accuracy from the HMMs 89\% to 97.7\%, significantly narrowing the gap between the disambiguator and the oracle.

Our second research question examines the impact of the improved lemma disambiguation accuracy on a downstream information retrieval (IR) task. 
Due to the lack of suitable Estonian datasets, we first translate the English DBpedia-entity dataset~\cite{Hasibi} into Estonian, employing the NLLB translation model~\cite{costa2022no}. We compare the performance of stemming, Vabamorf HMM-disambiguated lemmatization, and Vabamorf GliNER-disambiguated lemmatization in a BM25 retrieval setup. The results indicate ca 10\% improvement in retrieval metrics when using Vabamorf lemmatization over stemming, with an additional 1\% gain achieved through GliNER-enhanced disambiguation.

Overall, our contributions in this paper are threefold:

\begin{enumerate}
    \item We implement a new neural disambiguator based on an open-vocabulary span-labeling method for the Estonian rule-based morphological analyzer Vabamorf (henceforth referred to as GliLem) and show that it considerably improves the lemmatization results over the existing HMM-based disambiguator.
    \item We produce and release the first IR dataset for Estonian by machine translating the English  DBpedia-entity dataset.\footnote{\url{https://huggingface.co/datasets/adorkin/dbpedia-entity-est}}
    \item We demonstrate the efficacy of the proper lemmatization over stemming for the IR task in Estonian, showing also that improved disambiguation translates into up to 1\% improvement in the IR metrics.
\end{enumerate}

\section{ Vabamorf and GliNER}

In this section, we first give an overview of both the Estonian morphological analyzer Vabamorf and the open-vocabulary NER model GliNER.

\subsection{Vabamorf}

Vabamorf \cite{kaalep2001complete}
is a comprehensive, rule-based morphological analyzer specifically developed for the Estonian language. 
It leverages extensive morphological rules to generate all possible morphological analyses, including lemma candidates, for each analyzed word token. 
The analyzer accounts for the rich inflectional patterns of Estonian, which include numerous cases, tenses, and degrees of comparison.

Because many Estonian words can have several morphological analyses, Vabamorf includes a built-in HMM-based disambiguator, which aims to rank these candidates based on the contextual likelihood.
However, under the HMM formulation, the disambiguation context is very limited, with the analysis of the current word only being dependent on the analysis of the previous word. Therefore, the performance of the HMM-based disambiguator is more than 10\% lower than the oracle accuracy that can be obtained on the Estonian UD dataset \cite{dorkin-sirts-2023-comparison}. We used Vabamorf via EstNLTK, which is a library that provides an API to various Estonian language technology tools \cite{ORASMAA16.332}.

\subsection{GliNER}

\begin{figure*}[t]
  \centering
  \includegraphics[width=0.95\textwidth]{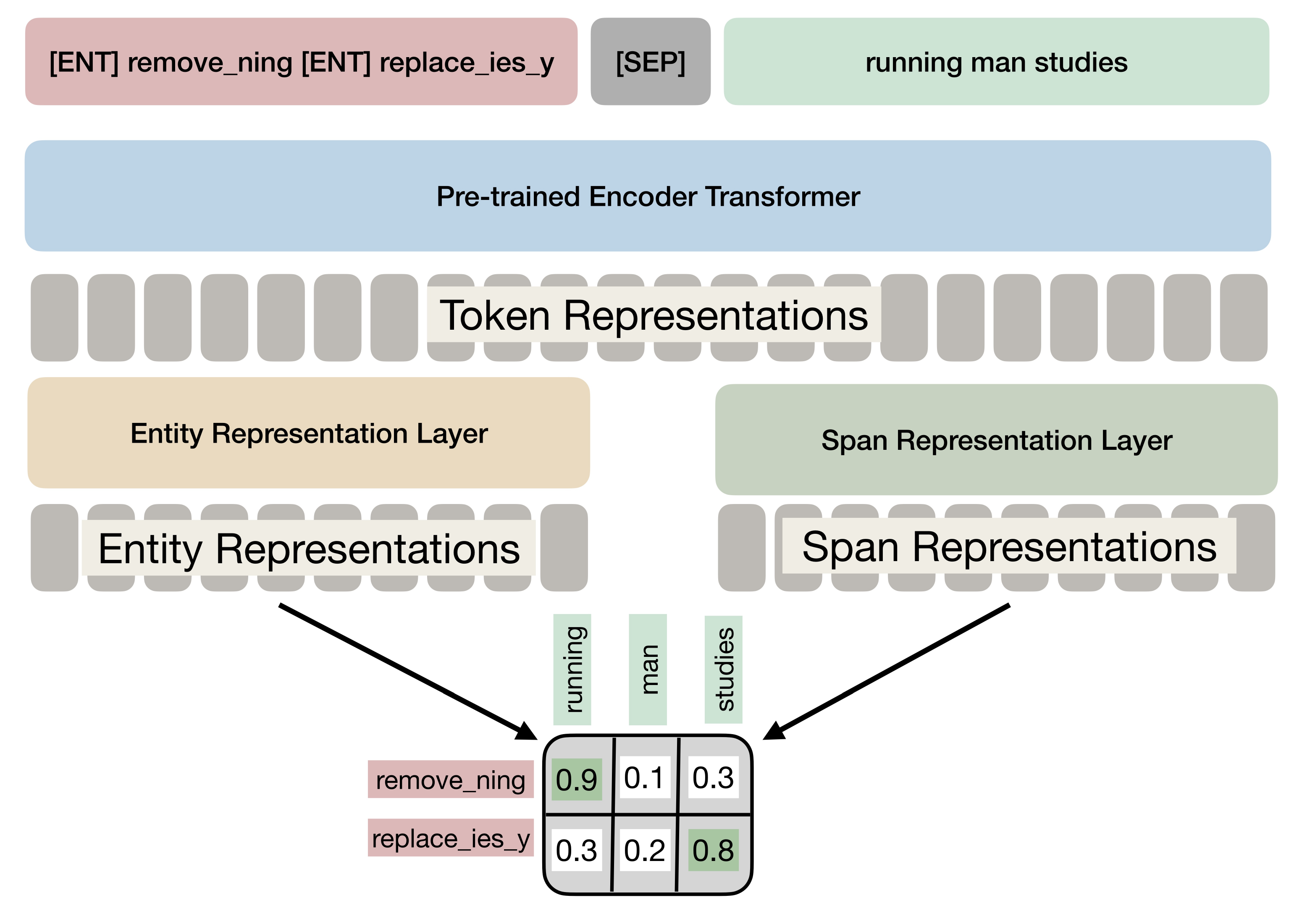}
  \caption{Schematic representation of the GliNER architecture applied to lemmatization.
  }
  \label{fig:arch}
\end{figure*}

GliNER~\cite{zaratiana2023gliner} is an open-vocabulary Named Entity Recognition (NER) model that extends traditional NER capabilities by allowing the labels to be specified in natural language (as opposed to nominal labels represented as integer indices in traditional classification models). Unlike conventional NER models that rely on a fixed set of entity types, GliNER can handle an arbitrary number of labels, making it highly adaptable for tasks requiring flexible label sets. 

GliNER is based on an encoder-only BERT-like architecture, which is expanded with span representation and entity representation modules (see Figure~\ref{fig:arch}). The modules are used to produce span and entity embeddings, accordingly. Span and entity embeddings are then used to measure pairwise similarity to identify entities in the input text. 
Entity types are expressed in natural language and separated from each other with the special \textbf{[ENT]} token. The entity types and input text are separated from each other with a \textbf{[SEP]} token, and they are processed in the model simultaneously in a cross-encoder fashion.

To implement GliNER, \citet{zaratiana2023gliner} take an existing pre-trained encoder model as a basis for both the span and entity representation modules, and add two blocks of feed-forward layers on top of the encoder to process the spans and entities separately. Finally, entities are assigned to spans by scoring the similarities between the output representations from both the span and entity modules.

GliNER was pretrained on Pile-NER\footnote{\url{https://huggingface.co/datasets/Universal-NER/Pile-NER-type}}~\cite{zhou2023universalner}, which is a synthetically annotated large scale NER dataset derived from the Pile corpus \cite{Gao2020ThePA} that has ca 13K distinct entity types. Such pretraining is expected to give the GliNER model an ability to generalize to very different types of labels.

\section{Adapting GliNER for Vabamorf Lemma Disambiguator}

We observe that the GliNER architecture is flexible enough to be used for essentially any kind of token classification task, including part-of-speech tagging and morphological analysis. 
To be applicable for lemmatization, we adopt the approach proposed by~\citet{straka-2018-udpipe} that expresses each example of \textit{form} $\rightarrow$ \textit{lemma} as a transformation rule. Each transformation rule comprises the minimal sequence of character-level edits---commonly referred to as a shortest edit script---such as adding, removing, or replacing characters, required to transform a given \textit{form} into its \textit{lemma}.
The transformation rules are represented simply as string labels, which are then used in token classification. For specific examples of transformation rules refer to Table~\ref{tab:dist}.

While in theory it would be possible to use lemmas directly as ``entities'' to be scored instead of transformation rules, that would inflate the number of ``entity types'' to be learned considerably. Effectively, each token type would have to have its own lemmatization label. Meanwhile, the transformation rules proposed by~\citet{straka-2018-udpipe} are abstract enough to allow for compact representation of similar transformations, and, according to~\citet{toporkov2024evaluating}, they offer stronger generalization than alternative approaches to shortest edit script generation. For instance, some common rules, such as ``do nothing'' and ``upper case the first character'', are easily applicable to any surface form.

The total number of unique transformation rules relevant for Estonian is too large to be used as input to GliLem. However, we only aim to score and rank the lemma candidates that the rule-based Vabamorf outputs for each token in the text. This limits the total number of possible ``entities'' to be scored by the number of tokens in the text, but usually it is much lower than that, mainly because the ``do nothing'' rule is the most common by far even in morphologically rich Estonian (see Table~\ref{tab:dist}). According to \citet{toporkov2024role}, the ``do nothing'' rule is also the most common rule in diverse languages such as Basque, English, Russian, and Spanish. In Estonian, major contributors to the frequency of this rule, in addition to punctuation marks, include conjunctions, adverbs, some types of adjectives, postpositions, and inflected forms homonymous with the base form.

For each input token Vabamorf outputs at least one morphological analysis (together with lemma). Accordingly, to prepare Vabamorf outputs for disambiguation, a transformation rule is found for each token and all of its lemma candidates. 
The set of strings representing the obtained unique transformation rules are given as entities in the GliLem input. 
GliLem outputs a list of spans, each accompanied by a proposed matching transformation rule and its score. 
The obtained rules are then applied to the respective spans to get the lemmas.
The overall GliLem architecture, i.e., the GliNER architecture applied to lemmatization disambiguation, is schematically represented in Figure~\ref{fig:arch}.

\section{Enhancing Disambiguation with GliLem}

We implement the GliLem for disambiguating lemma candidates generated by Vabamorf.
To assess the effectiveness of the GliLem approach we evaluate the following approaches:

\begin{enumerate}
    \item Vabamorf lemmatization using the built-in HMM-based disambiguator;
    \item Vabamorf lemmatization in the Oracle mode (the prediction is considered correct if the correct lemma is in the proposed non-disambiguated candidates);
    \item Pattern-based token classification model for lemmatization;
    \item Vabamorf lemma candidates disambiguated with GliLem.
\end{enumerate}

\begin{table}[t]
    \centering
    \small
    \begin{tabular}{lll}
    \toprule
    \% & Rule & Description \\
    \midrule
    49.6 & \texttt{↓0;d¦} & Do nothing \\
    7.0 & \texttt{↓0;d¦-} & Remove the last letter \\
    4.8 & \texttt{↓0;d¦---} & Remove two last letters \\
    3.7 & \texttt{↑0¦↓1;d¦} & Upper case the first letter \\
    3.3 & \texttt{↓0;d¦-+m+a} & Replace the last letter with \textit{ma} \\
    3.2 & \texttt{↓0;d¦----} & Remove three last letters \\
    \bottomrule
    \end{tabular}
    \caption{Top 6 most common transformation rules present in the train split of the EDT dataset.
    }
    \label{tab:dist}
\end{table}

\subsection{GliLem training}

We conducted experiments using the Estonian Universal Dependencies EDT corpus version 2.14, using the pre-defined splits. During training we do not make use of Vabamorf. Instead, we convert the token/lemma pairs provided in the corpus into respective lemmatization labels (transformation rules) and format the data according to the GliNER convention.

GliNER annotation schema differs from the BIO scheme typically used in the NER task. In GliNER, entire token spans with corresponding labels are used as inputs, 
and more importantly for our case, non-entities, i.e., the most common ``default'' class, are not labeled. Correspondingly, we do not use the ``do nothing'' rule as a label, and instead consider it the default state of the token. That means, we only score cases where the lemma is different from the surface form.

For training the GliLem, we use the GliNER training script provided by the authors\footnote{\url{https://github.com/urchade/GLiNER/blob/main/train.py}} using the default parameters and our lemmatization data to train the multilingual version of the pretrained model. The reason for using this model as a base model instead of initializing span and entity modules from scratch is that we expect to benefit from multilingualism of the backbone encoder, and also from the learned span representations of the NER model itself.

\subsection{Token Classification Baseline}

To contextualize the effect of Vabamorf disambiguation with GliLem, we reproduce the experiments by \citet{dorkin-sirts-2023-comparison} with some differences. We reduce the amount of preprocessing applied to the dataset: we do not lowercase the data and do not remove the derivational symbols present in some lemmas. We also use the more recent UD version (2.10 $\rightarrow$ 2.14).

The token classification model is a simple, efficient, and computationally cheap baseline to offset the complexity of the GliNER-based approach. For that reason we do not directly reproduce the token classification approach of  \citet{dorkin-sirts-2023-comparison}, but rather use the adapter-based parameter efficient fine-tuning \cite{houlsby2019parameter}, which reduces the training time down to minutes.

We do not reproduce the results of the generative character-level transformer model~\cite{DBLP:journals/corr/abs-2005-10213} that \citet {dorkin-sirts-2023-comparison} reported as the highest scoring approach, because it requires additional morphological annotation as input. Essentially, it needs the data to be disambiguated first which is contradictory to our goals in this work.

\subsection{Results}

\begin{table*}[ht]
\centering
\begin{tabular}{l cc}
\toprule
\textbf{Method} & \textbf{Dev} & \textbf{Test} \\
\midrule
Vabamorf & 0.878 [0.877, 0.883] & 0.892 [0.889, 0.895] \\
Oracle Vabamorf & 0.992 [0.992, 0.993] & 0.993 [0.992, 0.994] \\
Pattern-based Token Classification & 0.962 [0.960, 0.964] & 0.966 [0.964, 0.968] \\
GliLem & \textbf{0.974 [0.973, 0.976]}  & \textbf{0.977 [0.975, 0.978]}\\
\bottomrule
\end{tabular}
\caption{Bootstrap estimates of the lemmatization accuracy on the Estonian UD EDT dev and test sets with 95\% confidence intervals. Oracle Vabamorf considers the prediction correct if the correct lemma appears in non-disambiguated Vabamorf predictions, thus making it unusable in a practical scenario where no labels are available.}
\label{tab:edt}
\end{table*}

The lemmatization results are shown in Table~\ref{tab:edt}.
The GliLem model achieves the lemmatization accuracy of 97.7\% on the test, which significantly outperforms Vabamorf's disambiguator that scores only 89.2\% on the same set, demonstrating the efficiency of a more advanced disambiguation approach.

The pattern-based token classification model that does not utilize Vabamorf's candidates reached 96.2\% accuracy. While the difference with the GliLem disambiguation is modest (only ca 1.2\% in absolute), it suggests that leveraging Vabamorf's morphological analysis combined with GliLem's disambiguation capabilities provides a performance advantage. Moreover, lemmatization accuracy scores are skewed towards higher values due to the majority of corpus tokens requiring no changes to transform the initial word form into the lemma, and that is generally not very difficult for any model to learn and predict.

In the Oracle mode, Vabamorf achieves an accuracy over 99\%, showing that the disambiguator module has still room for improvement. However, the gap with the GliLem is less than 2\% in absolute that can be hard to close.

\section{Impact on Information Retrieval}

The problem of lemmatization is usually evaluated in isolation, separately from an actual application. 
While lemmatization can be a useful step in some realistic scenarios, the impact of the improvement in the lemmatization accuracy on the improvement of the downstream task can be difficult to estimate.
To emulate the realistic scenario, we evaluate both the original Vabamorf disambiguator and the GliLem disambiguator in an information retrieval (IR) task.
While the IR task is nowadays often addressed with dense vector retrieval, hybrid methods that, as a first step, adopt lexical search methods are still highly relevant. 
Input normalization via lemmatization is also more important in morphologically complex languages that typically have less resources than English.
In particular, there is currently no IR benchmark dataset available in Estonian that would allow to evaluate the effect of different text normalization methods to the IR task. The only previous work in Estonian related to information retrieval that we are aware of is by \citet{dorkin-sirts-2024-sonajaht}. However, this work addressed the problem of retrieving dictionary words based on their definitions using dense IR methods and did not deploy hybrid methods necessitating lexical normalization in the first steps.
For this reason, we first translate an existing English information retrieval dataset to the Estonian language.

\vspace{0.5cm}
\subsection{Dataset Preparation and Translation}

DBpedia-Entity v2~\cite{Hasibi} is a test collection for entity search evaluation, consisting of 467 queries with graded relevance judgments for entities from the DBpedia 2015-10 dump. In this work, we refer to the test collection together with the DBpedia dump as DBpedia-Entity. The collection comprises several distinct types of queries:

\begin{enumerate}
    \item Short, ambiguous queries searching for one particular entity;
    \item Information retrieval-style keyword queries;
    \item Queries seeking a list of entities;
    \item Natural language questions answerable by DBpedia entities.
\end{enumerate}

\vspace{0.5cm}

For each query there is a list of a variable number of documents and their relevance judgments: highly relevant (2), relevant (1), irrelevant (0).
Each document in the corpus represents an entity which has an ID, a title in natural language, and a variable length description.
The dataset corpus---DBpedia 2015-10 dump---comprises approximately 4.5 million documents. 
We chose this dataset due to its general domain, the variety of query types it contains, and its focus on retrieving information from a very large collection of documents. 

To evaluate the effect of lemmatization accuracy on information retrieval quality in Estonian, we translated the DBpedia-Entity dataset into Estonian using the NLLB~\cite{costa2022no} translation model. 
We translated both documents and queries using the NLLB 3B,\footnote{\url{https://huggingface.co/facebook/nllb-200-3.3B}} which is the largest available dense version of the NLLB. We adopted the CTranslate2\footnote{\url{https://github.com/OpenNMT/CTranslate2}} library for efficient translation at large scale. Translating the entire dataset took approximately two days on a single A100 GPU on the University's High Performance Cluster~\cite{https://doi.org/10.23673/ph6n-0144}.

At this time, we did not perform any quantitative quality evaluation of the resulting translations. Based on the manual examination of a small sample of examples, we note that while the translation quality is far from perfect, it generally preserves the meaning well enough to be useful for our benchmark.

\subsection{Retrieval Experiments}

\begin{table*}[h]
\centering
\begin{tabular}{lccccc}
\toprule
\textbf{Metric} & \textbf{Baseline} & \textbf{Stemming} & \textbf{Vabamorf} & \textbf{GliLem} \\
\midrule
Recall@1       & 0.0269 & 0.0260 & 0.0218 & \textbf{0.0278} \\
Recall@5       & 0.0633 & 0.0627 & 0.0702 & \textbf{0.0734} \\
Recall@100     & 0.2212 & 0.2167 & 0.2831 & \textbf{0.2935} \\
\midrule
MAP@1          & 0.2077 & 0.2120 & 0.2527  & \textbf{0.2591} \\
MAP@5          & 0.1201 & 0.1312 & \textbf{0.1596}  & 0.1577 \\
MAP@100        & 0.0874 & 0.0856 & 0.1057 & \textbf{0.1115} \\
\midrule
Success@1      & 0.2077 & 0.2120 & 0.2527  & \textbf{0.2591} \\
Success@5      & 0.3704 & 0.4004 & \textbf{0.4925}  & 0.4797 \\
Success@100    & 0.6681 & 0.6767 & \textbf{0.7901} &  0.7837 \\
\bottomrule
\end{tabular}
\caption{Retrieval metrics for the proposed token normalization approaches on the translated DBpedia-Entity dataset.}
\label{tab:retrieval}
\end{table*}

The BM25 algorithm~\cite{robertson1995okapi} is considered a strong information retrieval baseline to this day even when compared to modern dense retrieval models~\cite{karpukhin2020dense, thakur2021beir}. BM25 relies on sparse lexical representations of documents and queries, with word-level tokens most commonly used for these representations. The tokens usually undergo additional preprocessing to account for surface form variation. For example, in sparse lexical representation, the present simple and the present participle forms of the word ``run'' (``run'' and ``running'') are considered entirely unrelated. That makes it difficult for the user to formulate queries because they have to guess in what form the desired term appears in the documents. For English, applying a stemming algorithm such as PorterStemmer is generally sufficient to deal with this problem. 

Meanwhile, stemming algorithms do not perform well for morphologically rich languages like Estonian due to significant variation in stem surface forms in many words. This scenario highlights a practical application of lemmatization---improving the quality of lexical search in such languages. While it is intuitive to expect that lemmatization can help, there are no previous works showing that for the Estonian language. Moreover, it needs to be clarified what effect the additional lemmatization accuracy obtained from better disambiguation of Vabamorf outputs has on information retrieval.

For our experiments we used the recent BM25s library\footnote{\url{https://github.com/xhluca/bm25s}} \cite{bm25s} that provides a fast implementation of BM25. For indexing, we used the default parameters and omitted the preprocessing done by the library---we input the corpus preprocessed by us directly.

We preprocessed the Estonian translation of the DBpedia-Entity corpus by applying the following four preprocessing approaches to the dataset documents:\footnote{We exclude the token-classification baseline because we are interested in gauging the effect of improved lemmatization disambiguation on IR specifically.}

\begin{enumerate}
    \item Identity (only tokenization is applied);
    \item Stemming using the Estonian Stemmer available in Apache Lucene;\footnote{\url{https://lucene.apache.org/core/8_11_0/analyzers-common/org/apache/lucene/analysis/et/EstonianAnalyzer.html}}
    \item Vabamorf lemmatization with the built-in HMM disambiguation;
    \item Vabamorf lemmatization with the GliLem disambiguation.
\end{enumerate}

The output from each preprocessing resulted in each document being represented as a list of tokens, which were then concatenated with whitespace, the expected input format for BM25. The entire corpus was then passed to the indexer implementation. The indexing process took about three minutes, regardless of the preprocessing type.

Finally, we applied the same preprocessing options to the translated queries and, for each query, retrieved \textbf{100} most relevant documents from the corpus. Then, we employed relevance judgments from the original DBpedia-entity dataset to obtain the ground truth documents for each query (we selected only the documents deemed relevant or highly relevant) to calculate several retrieval metrics explained in the next section.

\subsection{Evaluation Metrics}

\textbf{Success@k} measures whether a user's information need is satisfied by at least one result in the top \textbf{k} retrieved items~\cite{karpukhin2020dense, khattab2021relevance}. It is a coarse-grained metric that does not distinguish how well the user's information need was satisfied.
\\

\noindent\textbf{Recall@k} measures what percentage of all relevant items for a query appear within the top \textbf{k} retrieved results~\cite{buttcher2016information}. The metric is suitable for our case, because only a small proportion of the total number of documents is annotated with relevance judgments and therefore the Recall will be upper bounded only with very small k values.\footnote{Consider for instance the case where there are 1000 relevant documents per query. In this case, for instance with k of 100, the Recall will be upper bounded by 0.1.} 
\\

\noindent\textbf{Mean Average Precision (MAP)@k} measures both the precision and ranking quality of the results up to position \textbf{k}, averaged across all queries. It captures not just whether relevant items were retrieved, but also how high they were ranked, giving more weight to relevant items appearing higher in the results~\cite{buttcher2016information}. This metric prioritizes results that group relevant results closer to the top.

\subsection{Results and Discussion}

The IR performance measures at several \textbf{k}-s are shown in Table~\ref{tab:retrieval}.
First, we observe that the baseline of using word forms is on the same level with stemming on all measures, which is due to the Apache Lucene stemmer, although Estonian-specific, being very weak for Estonian.

When looking at the setting with \textbf{k} equal to 1, the Recall does not change considerably, but both MAP and Success rate (that are by definition equal in this setting) improve more than 4\%  when using lemmatization over stemming, although enhanced disambiguation with GliLem gives only a minor improvement over the default Vabamorf disambiguation.

In the \textbf{k} equal to 5 setting, Recall improves about 1\%, MAP about 3\%, and the Success rate, which is the most lenient measure, improves about 9\%, when comparing Stemming to lemmatization with Vabamorf. In this setting, only the Recall measure shows a small positive impact of the more complex disambiguation with GliLem over the default Vabamorf disambiguation, while for the MAP and Success rate, the baseline Vabamorf gives better results.

Finally, in the \textbf{k} equal to 100 setting, when comparing lemmatization to stemming, Recall improves ca 7\%, MAP about 2\% and Success rate improves about 11\%, with GliLem disambiguation showing an additional improvement of ca 1\% in both Recall and MAP over the Vabamorf default disambiguation.

We conclude that proper lemmatization can considerably improve IR results compared to stemming. At the same time, even large improvements in lemmatization accuracy, obtained by replacing the simple HMM-based disambiguation component with a more complex GliNER-based disambiguation do not easily translate into significantly better IR results.
However, when comparing the baseline Vabamorf with the GliLem disambiguation, we observe a small but consistent improvements in Recall for all values of \textbf{k}, with the improvement being the most pronounced in the highest \textbf{k} setting.
Using a high \textbf{k} is typical in hybrid IR systems, where the lexical retrieval is the first step to reduce the number of potentially relevant documents. Thus, the relatively small lemmatization improvement can have a positive effect in the downstream IR task.

Upon manual inspection of the original DBpedia-Entity corpus, we observed that it is somewhat noisy. Some entries have little to no content, while others are comprised of large listings. Many entries have characters from diverse writing systems. This results in additional noise introduced during the imperfect translation process. Moreover, there are translation errors in the translated queries (such as the presence of non-existent words), as well. We believe that the positive effect of the improved lemmatization being somewhat small can be at least partially  attributed to these issues. Accordingly, some future work could be dedicated to improving the translated dataset, e.g.,  manually correcting the query translations, performing translation quality estimation to redo or filter out low quality document translations, and filtering out entries with no useful content. Consequently, we would expect a larger positive effect of improved lemmatization on a corrected dataset. However, we believe that the noisiness of the dataset affects each approach similarly, and thus the relative ranking between the preprocessing methods remains stable.

We also note that both disambiguation approaches are somewhat computationally intensive. In the current implementation of GliNER, the batch processing does not allow different sets of labels for each example, and thus each example must be processed separately, which makes it difficult to make use of GPU acceleration during inference. The Vabamorf disambiguator, on the other hand, cannot be accelerated at all. Applying both disambiguation approaches to the large corpus of 4.5M documents took over 50 hours for each, using parallelization with approximately 30 concurrent processes on CPU hardware.

\section{Conclusion}

This study demonstrates that integrating an external disambiguation model like GliLem with a rule-based morphological analyzer can substantially improve the accuracy of lemmatization in Estonian. The enhanced lemmatization bridges the accuracy gap caused by the limitations of Vabamorf's built-in disambiguator. This proves our initial hypothesis that the main weakness of Vabamorf is in fact its inability to correctly select the lemma candidate in context.

Additionally, we estimated the effect of improved lemmatization accuracy on an information retrieval downstream task.  Although the precise effect is difficult to estimate due to the noisiness of the original data and additional noise introduced by imperfect machine translation, we observed small consistent improvements in Recall, and especially in the setting with a high k, suggesting that improved lemmatization might translate into actual improvements in a hybrid information retrieval setting.

\section*{Acknowledgments}

This research was supported by the Estonian Research Council Grant PSG721.

~

\bibliographystyle{acl_natbib}
\bibliography{nodalida2025}

\end{document}